\documentclass[twoside,11pt]{article}

\usepackage[preprint]{jmlr2e}
\usepackage[utf8]{inputenc} 
\usepackage[T1]{fontenc}    
\usepackage{hyperref}       
\usepackage{url}   
\usepackage[ruled,longend,linesnumbered]{algorithm2e}
\usepackage{float}
\usepackage{multirow}
\usepackage{booktabs}       
\usepackage{amsfonts}    \usepackage{rotating}
\usepackage{nicefrac}       
\usepackage{microtype}  
\usepackage{graphicx}
\usepackage{tabularx}
\usepackage{subfigure}
\usepackage[dvipsnames]{xcolor}
\usepackage{graphicx}
\usepackage{amsmath}
\usepackage{xcolor}
\usepackage{colortbl}
\usepackage{enumitem}
\usepackage{caption}
\usepackage{listings}
\usepackage{appendix}
\usepackage{chngcntr}
\usepackage{minitoc}
\usepackage{siunitx}

\newcolumntype{C}{>{\centering\arraybackslash}X}

\hypersetup{
    colorlinks = true,
    allcolors = {Tan},
    linkbordercolor = {white},
}

\ShortHeadings{Automated Lesion Segmentation in Whole-Body PET/CT}{Jiayi Liu et al.}
\firstpageno{1}

\begin{document}

\doparttoc 
\faketableofcontents 

\title{Enhancing Lesion Segmentation in PET/CT Imaging with Deep Learning and Advanced Data Preprocessing Techniques}

\author{\name Jiayi~Liu \email jiayi.liu01@cri-united-imaging.com \\
    \addr Shanghai United Imaging Healthcare Advanced Technology \\ Research Institute Co., Ltd. \\
    Shanghai 201807, China
    \AND
    \name Qiaoyi~Xue  \email 
    qiaoyi.xue@cri-united-imaging.com \\
    \addr Shanghai United Imaging Healthcare Advanced Technology \\ Research Institute Co., Ltd. \\
    Shanghai 201807, China
    \AND
    \name Youdan~Feng \email youdan.feng@cri-united-imaging.com \\
    \addr Shanghai United Imaging Healthcare Advanced Technology \\ Research Institute Co., Ltd. \\
    Shanghai 201807, China
    \AND
    \name Tianming~Xu \email cecilia\textunderscore xtm@sjtu.edu.cn  \\
    \addr Global Institute of Future Technology \\ Shanghai Jiao Tong University \\
    Shanghai 200240, China
    \AND
    \name Kaixin~Shen \email 812852899@sjtu.edu.cn \\
    \addr Global Institute of Future Technology \\ Shanghai Jiao Tong University \\
    Shanghai 200240, China
    \AND
    \name Chuyun~Shen \email 
    cyshen@stu.ecnu.edu.cn \\
    \addr School of Computer Science and Technology \\
    East China Normal University\\
    Shanghai 200062, China
    \AND
    \name Yuhang~Shi \email yuhang.shi@cri-united-imaging.com \\
    \addr Shanghai United Imaging Healthcare Advanced Technology \\ Research Institute Co., Ltd. \\
    Shanghai 201807, China
}

\maketitle

\begin{abstract}
The escalating global cancer burden underscores the critical need for precise diagnostic tools in oncology. This research employs deep learning to enhance lesion segmentation in PET/CT imaging, utilizing a dataset of 900 whole-body FDG-PET/CT and 600 PSMA-PET/CT studies from the AutoPET challenge III. Our methodical approach includes robust preprocessing and data augmentation techniques to ensure model robustness and generalizability. We investigate the influence of non-zero normalization and modifications to the data augmentation pipeline, such as the introduction of RandGaussianSharpen and adjustments to the Gamma transform parameter. This study aims to contribute to the standardization of preprocessing and augmentation strategies in PET/CT imaging, potentially improving the diagnostic accuracy and the personalized management of cancer patients.
Our code will be open-sourced and available at \href{https://github.com/jiayiliu-pku/DC2024}{https://github.com/jiayiliu-pku/DC2024}.

\end{abstract}

\section{Introduction}\label{sec:intro}

\begin{figure}[htb!]
    \centering
    \includegraphics[width=0.9\linewidth]{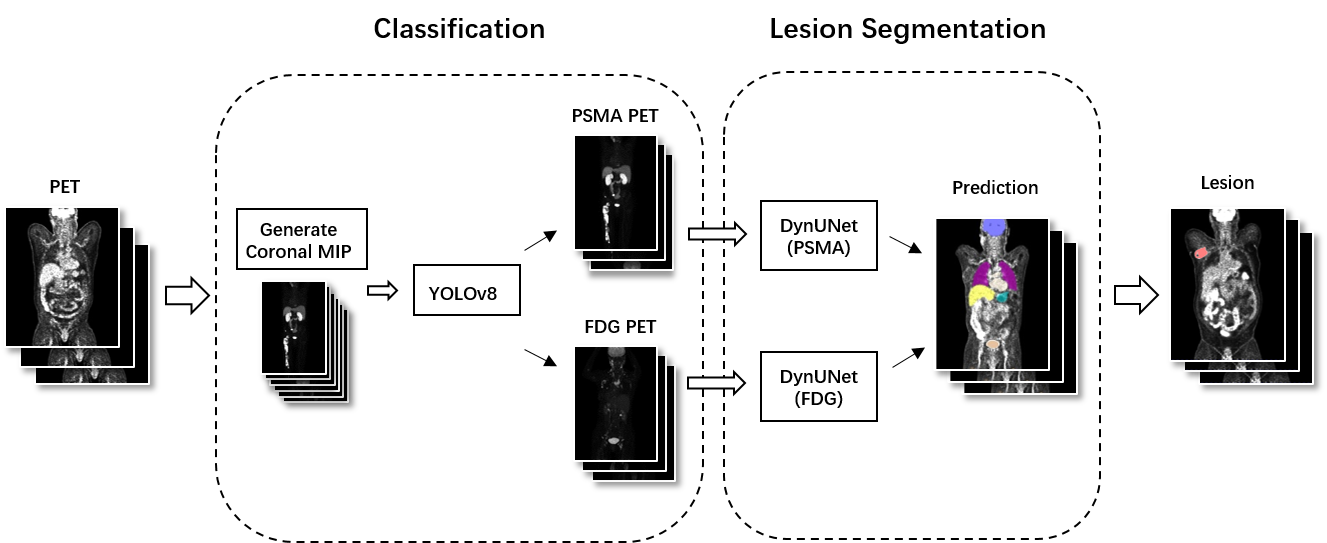}
    \caption{The workflow of automated lesion segmentation of FDG PET images and PSMA PET images.}
    \label{fig:fig1}
\end{figure}

The escalating incidence of cancer worldwide necessitates the advancement of diagnostic and therapeutic technologies that can enhance the precision and personalization of cancer care. The integration of diagnostic imaging with targeted therapy, as exemplified by molecular theranostics, has emerged as a pivotal strategy in delivering personalized treatment with enhanced accuracy. Within this domain, the conjunction of positron emission tomography (PET) and computed tomography (CT) plays a critical role in the diagnostic arsenal for oncological applications.
The deployment of radiotracers such as Fluorodeoxyglucose (FDG) and Prostate-Specific Membrane Antigen (PSMA) in PET/CT imaging has been instrumental in the effective detection and management of various cancer types. FDG PET/CT is particularly adept at highlighting metabolically active cancer cells, facilitating the evaluation of multiple cancer entities ~\citep{ref-1}. Similarly, the high expression of PSMA in prostate cancer cells underscores its diagnostic and therapeutic relevance, making it a valuable target for both imaging and therapeutic interventions ~\citep{ref-2,ref-3}.
The incorporation of deep learning (DL) techniques into PET/CT imaging has significantly enhanced the accuracy of lesion segmentation. DL models have facilitated the delineation of lesions in FDG PET/CT imaging, addressing the challenges associated with distinguishing pathological changes from physiological uptake in organs such as the liver and brain ~\citep{ref-4}. Advances in multi-label segmentation methodologies have further improved the precision of lesion delineation by enabling the concurrent identification of lesions and organs with high radiotracer uptake~\citep{ref-5,ref-6}. In the context of PSMA PET imaging, DL has become increasingly vital for the early detection of lymph node metastases and for monitoring therapeutic responses, demonstrating superior performance compared to traditional imaging modalities ~\citep{ref-7,ref-8}.
Despite these advancements, the scarcity of publicly available PET data presents a significant hurdle in the medical deep learning community, hindering the development of standardized preprocessing approaches for PET images, including normalization and augmentation techniques. This study aims to address this challenge by investigating and developing innovative methods for data preprocessing and postprocessing to enhance the accuracy and reliability of PET/CT lesion segmentation. Through rigorous academic inquiry and practical application, we seek to contribute to the armamentarium of clinical tools that can improve the diagnostic and therapeutic process for cancer patients.

\section{Methods}

\subsection{Data and preprocessing}

The training of the FDG models was conducted using whole-body FDG-PET/CT data from a cohort of 900 patients, encompassing 1014 studies supplied by the AutoPET challenge III in 2024. The challenge consists of patients with malignant melanoma, lymphoma, lung cancer and negative control patients. The data was split into a training set of 810 cases and a testing set of 204 cases. For the PSMA model, 600 PSMA-PET/CT data supplied by the AutoPET challenge III was split into a training set of 479 cases and a testing set of 121 cases. Lesion numbers and patient meta info were taken into consideration to ensure that both the training and testing subsets exhibited equitable distributions of lesion counts. 

The preprocessing steps includes resampling to achieve uniform spatial spacing and intensity normalization. AutoPET III provides a default robust data augmentation pipeline to enrich the training dataset, incorporating spatial and intensity transformations. Random cropping and affine transformations (translation, rotation, scaling) simulated spatial variations, while Gaussian noise, smoothing, and sharpening techniques accounted for image quality diversity. Intensity adjustments included random scaling and contrast variations, both inverted and non-inverted. Random flipping along spatial axes further increased data variability, aiming to enhance the model’s generalization capabilities. The details of default preprocessing steps and data augmentation pipline could be found at \href{https://github.com/ClinicalDataScience/datacentric-challenge.git}{https://github.com/ClinicalDataScience/datacentric-challenge.git}. We further tested influence of \textbf{non-zero normalization}, \textbf{ClipValMax=280} (clip intensity>280 for PET image). Whether adding \textbf{RandGaussianSharpen} to data augmentation pipline or change the parameter ($\gamma$)  of \textbf{Gamma transform} to 1-1.5 (defualt $\gamma=0.7-1.5$) could enhance segmentation results would be also evaluated.

\subsection{Model architecture and training}

The automated lesion segmentation process for FDG and PSMA PET images consists of two steps. First, a YOLOv8 model was trained for tracer classification of PET medical images. Second, two 3d Unets were trained independently with FDG or PSMA data for lesion segmentation.
 
\subsubsection{Yolo model}
Details of Yolo model training could be find at ~\citep{ref-9}.

\subsubsection{DynUNet model}
DynUNet model and training configuration were fixed in datacentric challenge in AutoPET III. Please refer to \href{https://github.com/ClinicalDataScience/datacentric-challenge.git}{https://github.com/ClinicalDataScience/datacentric-challenge.git}

\section{Results}

The outcomes of this investigation highlight the efficacy of various preprocessing and data augmentation strategies on the segmentation performance of FDG and PSMA tracers (shown in Table.\ref{tab:tab1}). For the FDG tracer, the default settings yielded a Dice coefficient of 63.19\%, indicating moderate segmentation precision. In contrast, the PSMA tracer under default conditions presented a Dice score of 32.07\%, with notable false positive volumes, suggesting that the baseline approach is suboptimal for this tracer.

Enhancements in segmentation were observed when Gaussian sharpening was applied as a data augmentation technique, resulting in a Dice score of 44.55\% for the PSMA tracer and 64.11\% for the FDG tracer, accompanied by a reduction in false positive volumes. The most significant improvement was achieved with a max clip value of 280 in preprocessing combined with Gaussian sharpening, which elevated the Dice score to 53.69\% and minimized false positive volume to \SI{5.55}{cm^3} and \SI{15.28}{cm^3}, respectively.

\begin{table}[h]
\caption{Performance of Different Preprocessing Steps and Data Augmentation Methods.}
\centering
\resizebox{1\textwidth}{!}{%
\begin{tabular}{@{}l|c|c|c|c|c@{}}
\toprule
\textbf{Tracer} &\textbf{Preprocessing Steps} &\textbf{Data Augmentation} &\textbf{Dice} &\textbf{FPvol} & \textbf{FNvol} \\ 
\midrule
FDG &Default &Default &63.19 &8.28 &7.07  \\
\textbf{FDG} &\textbf{Default} &\textbf{Default+GuassianSharpen} &\textbf{64.11} &\textbf{2.48} &\textbf{9.67}                      \\
\midrule
PSMA &Default &Default                               &32.07 &76.53  &12.04                      \\
PSMA &Default(non-zero normalization) &Default       &29.93 &101.70 &16.57                      \\
PSMA &Default &Default+GuassianSharpen               &44.55 &20.92  &12.14                      \\
PSMA &Default &Default+GammaTransform(1-1.5)         &46.31 &4.74   &24.61                      \\
\textbf{PSMA} &\textbf{Default+ClipValMax=280} &\textbf{Default+GuassianSharpen} &\textbf{53.69} &\textbf{5.55}   &\textbf{15.28} \\

\bottomrule
\end{tabular}}
\label{tab:tab1}
\end{table}

\section{Conclusion}

In this paper, we compared segmentation results among different preprocessing steps and data augmentation methods.
These findings confirm that the careful selection and integration of preprocessing steps and data augmentation methods are pivotal in refining the segmentation capabilities of tracer-based imaging models, underscoring the necessity for customized approaches to optimize model performance.

\clearpage
\newpage

\bibliography{main}

\end{document}